\documentclass[aip,apl,reprint]{revtex4-1}
	\usepackage{lineno}
	\modulolinenumbers[5]

	\usepackage{amsmath}
	\usepackage{amssymb}
	\usepackage{graphicx}%
	\usepackage{dcolumn}%
	\usepackage{bm}%
	\usepackage[colorlinks=true,citecolor=blue,allcolors=blue]{hyperref}

	\usepackage{algorithm}
	\usepackage{algpseudocode}
	\usepackage{slashbox}

	\hypersetup{breaklinks=true}

	\newcommand{\gv}{\boldsymbol}
	\newcommand{\uv}[1]{\hat{\mathbf{#1}}}
	\newcommand{\ugv}[1]{\hat{\boldsymbol{#1}}}

	\newcommand{\mat}[1]{\mathrm{#1}}
	\newcommand{\set}[1]{{#1}}
	\newcommand{\st}{\;\vert\;}

	\newcommand{\refsub}[2]{\hyperref[#1]{\ref{#1}{\bf #2}}}

  \newcommand{\figref}[2]{\hyperref[#1]{\figurename~\ref{#1}{#2}}}
  \newcommand{\figsref}[4]{\figuresname~\hyperref[#1]{\ref{#1}{#2}} and \hyperref[#3]{\ref{#3}{#4}}}

\begin{document}

\title{Seed-Point Based Geometric Partitioning of Nuclei Clumps}

    \author{James Kapaldo}
    \email[Email: ]{jkapaldo@nd.edu}

    \affiliation{Department of Physics, University of Notre Dame, Notre Dame, USA}

    \begin{abstract}
        When applying automatic analysis of fluorescence or histopathological images of cells, it is necessary to partition, or de-clump, partially overlapping cell nuclei. In this work, I describe a method of partitioning partially overlapping cell nuclei using a seed-point based geometric partitioning. The geometric partitioning creates two different types of cuts, cuts between two boundary vertices and cuts between one boundary vertex and a new vertex introduced to the boundary interior. The cuts are then ranked according to a scoring metric, and the highest scoring cuts are used. This method was tested on a set of 2420 clumps of nuclei and was found to produced better results than current popular analysis software.
    \end{abstract}

    \date[Submitted: ]{\today}
	\maketitle

%\linenumbers

\section{Introduction}
\begin{figure*}[!t]
    \centering
    \includegraphics[width=\textwidth]{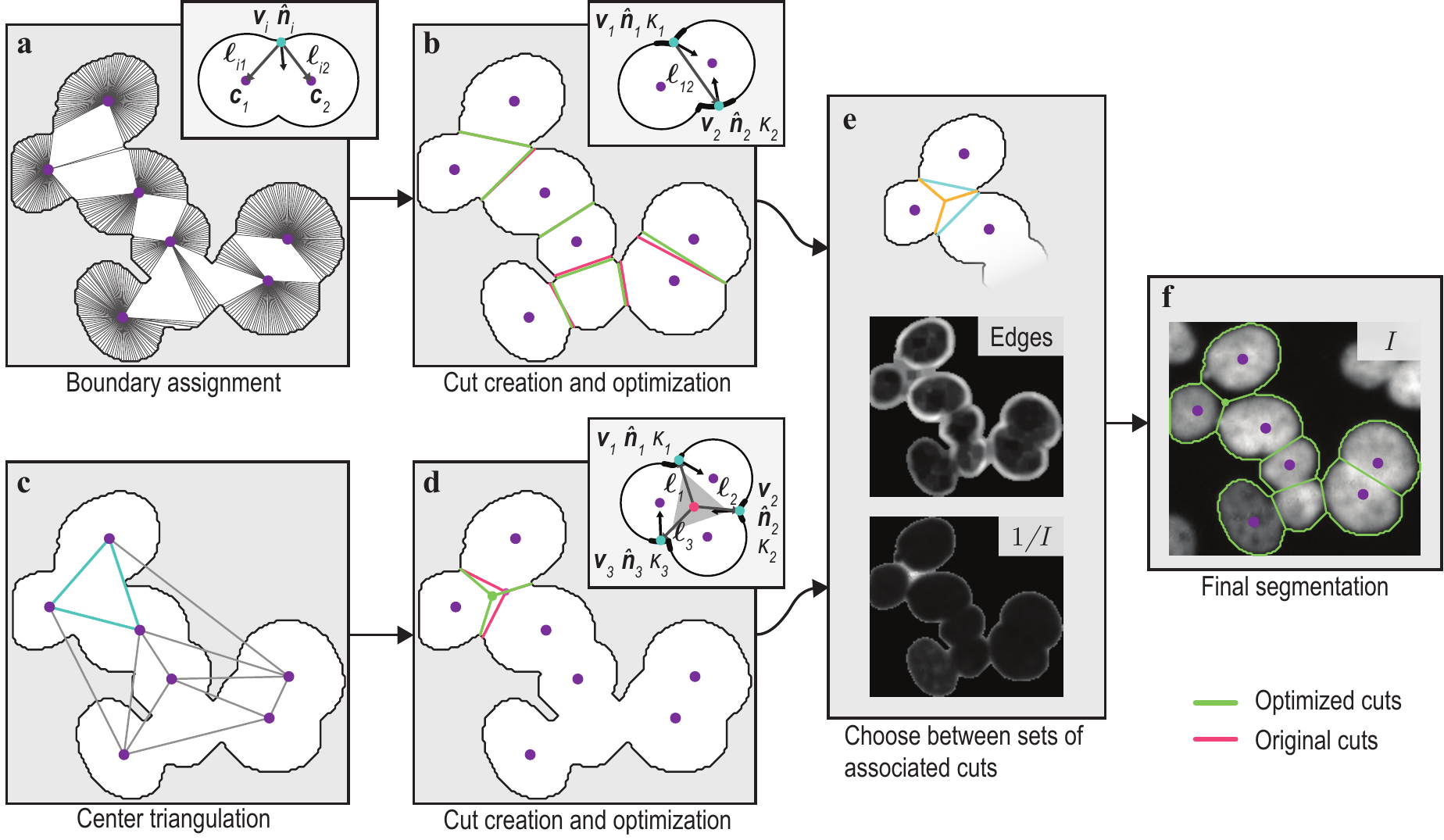}
    \caption{Geometric partitioning of nuclei clump. Vertex-vertex cuts: {\bf a}, assign boundary vertices to centers by optimizing directions and distances; {\bf b}, create cuts (red lines) using the set of vertices assigned to each center (while ignoring far away vertices) and then optimize the cuts (green lines) over a small range (bold region in inset) using direction, distance, and curvature. Vertex-center cuts: {\bf c}, create a Delaunay triangulation of the centers and remove triangles with edges that intersect the boundary or that have interior angles too small or too large; {\bf d}, create cuts (red lines) between the triangle center and boundary vertices nearest to the triangle's edge midpoints and then optimize the cut vertex locations over a small range (bold region in inset) using direction, distance, and curvature and optimize the triangle center location over a small region (shaded triangle in inset) using mean distance and direction. Choose between vertex-vertex and vertex-center cuts: {\bf e}, find sets of associated cuts and vote for the best set using direction, curvature, overlap with image edges, and overlap with one over the image intensity. {\bf f} shows the final segmentation. The insets of {\bf a}, {\bf b}, and {\bf d} depict the geometries for those parts where $\vec{v}_i$ is a boundary vertex, $\uv{n}_i$ is the inward pointing unit normal of the boundary at $\vec{v}_i$, $\kappa_i$ is the boundary curvature at $\vec{v}_i$, $\vec{c}$ is a nuclei center, and $\gv{\ell}$ is a displacement vector. (One of the nuclei centers is purposefully missing for demonstrative purposes.)\label{fig:nucleiPartitioning}}
  \end{figure*}

  When applying automatic analysis of fluorescence or histopathological images of cells, it is necessary to partition, or de-clump, partially overlapping cell nuclei. In this work, I describe a method of partitioning clumped nuclei using a seed-point based geometric partitioning. The method is summarized in \figref{fig:nucleiPartitioning}{} and described in the following sections.

  I assume that the nuclei centers (seed-points) have already been computed using, for instance, SALR clustering\cite{2018arXiv180404071K}, multi-pass voting\cite{Parvin2007}, single-pass voting\cite{Qi2012}, or other methods. Using these seed-points, I develop a simple rule based geometric method that creates two types of cuts, cuts between two boundary vertices (vertex-vertex) and cuts between a boundary vertex and a new vertex added to the interior region of the boundary (vertex-center), and the type of cut used in each region of the nuclei clump is decided by accumulating votes based on curvature, cut distance, overlap with image edges, and overlap with the inverted image.\par

  \section{Vertex-vertex cuts}
    \subsection{Boundary assignment}
      Each vertex of the boundary is assigned to a center (seed-point), see \figref{fig:nucleiPartitioning}{a}, by minimizing the distance between the boundary vertex and the center, by maximizing the dot-product between the boundary vertex normal and the assignment line pointing to the center, and by ensuring that no assignment lines overlap each other or the boundary. Mathematically, let $\vec{v}_i$ be the location of the $i$th boundary vertex, $\uv{n}_i$ be the inward pointing boundary unit normal vector at $\vec{v}_i$, $\vec{c}_j$ be the location of the $j$th center, and $\gv{\ell}_{ij}=\vec{c}_j-\vec{v}_i$ be an assignment vector from $\vec{v}_i$ to $\vec{c}_j$. (These definitions are shown in the inset of \figref{fig:nucleiPartitioning}{a}.) Give each possible assignment a score $\mat{S}_{ij}=(\ugv{\ell}_{ij}\cdot \uv{n}_i)/|\gv{\ell}_{ij}|$, and set any assignment with $\ugv{\ell}_{ij}\cdot \uv{n}_i < \theta_\mathrm{min}$ or $|\gv{\ell}_{ij}| > R_\mathrm{max}$ to be an invalid assignment, where $\theta_\mathrm{max}$ and $R_\mathrm{max}$ are thresholds that do not have a significant effect on the result, but can decrease computation time by removing improbable assignment options.
      \begin{enumerate}
        \item  \emph{Best assignment.} Assign each vertex to the center that maximizes the score. Save the assignment $a_i = \arg\min_j{(\mat{S}_{ij})}$, the score $s_i = \min_j{(\mat{S}_{ij})}$ and assignment vector $\gv{\ell}_i=\gv{\ell}_{ia_i}$.
        \item \emph{Removal of intersecting assignment vectors.} It may be that the assignment vectors intersect with each other. These intersections will lead to invalid partitioning of the clump and must be removed. For each boundary assignment vector $\gv{\ell}_i$, find the set of assignment vector indices $\set{K}_i$ that intersect with it and create a new score $\mathfrak{s}_i = \sum_{k\in\set{K}_i} s_{i}/s_{k}$. Remove the assignment vector with the smallest score $\mathfrak{s}$. Iterate this step until there are no more intersections.
        \item \emph{Finish.} If there were no intersections in step 2., stop. If there were intersections, then for each assignment vector that was removed in step 2., set that assignment in $\mat{S}_{ij}$ to invalid, and go back to step 1.
      \end{enumerate}
      Assignments that result from this algorithm are shown in \figref{fig:nucleiPartitioning}{a}. Note that this algorithm does not require all boundary vertices to be assigned to a center; this will be shown to be beneficial later.\par

      The threshold on assignment distance $R_\mathrm{max}$ should be set by considering the maximum single nuclei radius. The threshold on the angle $\theta_\mathrm{min}$ removes assignments where $\ugv{\ell}_{ij}$ and $\vec{n}_i$ do not point in the same direction; increasing this value can be helpful in successful partitioning of nuclei clumps even when centers are missing, but setting to a value too large ($\gtrsim0.8$) can have negative consequences, particularly when the nuclei have an abnormal shape, e.i.\ long and narrow. For all results shown in this paper, I used $R_\mathrm{max} = 35$ pixels and $\theta_\mathrm{min}=0.5$. \par

   \subsection{Cut creation}
      The boundary is broken up into pieces, where each piece is a set of adjacent vertices (without any breaks) all assigned to the same center. All of the pieces assigned to the same center are collected into one set, and then ordered so that the pieces form a well oriented boundary as follows. Consider the set of pieces assigned to the $i$'th center, let $\vec{v}_{s;k}$ be the first vertex (start) of the $k$'th piece in the set, and let $\vec{v}_{e;k}$ be the last vertex of that piece; further, let $\uv{n}_{(s,e);k}$ denote the normal vectors at these vertices. Define the first piece $k_1$ in the well oriented boundary to be the piece closest to the center, the second piece $k_2$ will be
      \begin{equation}
      k_2 = \arg\max_m \frac{\uv{n}_{e;k_1}\cdot\ugv{\ell}_{mk_1} - \uv{n}_{s;m}\cdot\ugv{\ell}_{mk_1}}{|\gv{\ell}_{mk_1}|}
      \end{equation}
      where $\gv{\ell}_{mk}=\vec{v}_{s;m}-\vec{v}_{e;k}$. From the second piece, the third piece is found, and so on until we return to a piece that has already been visited. Any piece not visited is ignored. Using the now ordered/oriented pieces, a cut is created whenever the distance between two adjacent vertices in a set is larger than one pixel. These cuts are shown as the red lines in \figref{fig:nucleiPartitioning}{b}.\par

      Allowing for vertices to be unassigned in the boundary assignment together with ignoring pieces that are far away and do not have high normal overlap allows for proper partitioning of a nuclei clump even when one of the nuclei's centers is missing, which is demonstrated by the missing center and proper partitioning in the bottom of the nuclei clump shown in \figref{fig:nucleiPartitioning}{}.

     \subsection{Cut optimization}\par
      The vertices of the cuts found above are not necessarily in the optimal position in their local neighborhood; thus, each cut is optimized by looking for a new vertex near each cut vertex that minimizes the cut distance, maximizes the boundary curvature at each vertex, and maximizes the dot-product between the cut vertex boundary normals and the cut line. In detail, let $\vec{v}_{1,2}$, $\uv{n}_{1,2}$, and $\kappa_{1,2}$ be the two vertices, boundary unit normals, and boundary curvatures of a cut, and let the sets of boundary indices in small neighboring regions around each vertex $\vec{v}_{1,2}$ be denoted $\set{V}_{1,2}$. (These definitions and the small regions, denoted by bold boundaries, are shown in the inset of \figref{fig:nucleiPartitioning}{b}.) The optimization searches for two new vertices $\vec{v}_p,\vec{v}_q$ such that
      \begin{equation}
      \label{eq:vv_cutOptimization}
      \{p,q\} = \underset{\{i\in\set{V}_1,j\in\set{V}_2\}}{\arg\max} \frac{\uv{n}_i\cdot\ugv{\ell}_{ji}-\uv{n}_j\cdot\ugv{\ell}_{ji}+\kappa_i+\kappa_j}{|\gv{\ell}_{ji}|}
      \end{equation}
      and where $\gv{\ell}_{ji} = \vec{v}_j - \vec{v}_i$. In fact, before using \eqref{eq:vv_cutOptimization}, all negative curvatures are first multiplied by five to help ensure that cut vertices are more likely at concave locations. The radius of the small region around the vertices used in all results shown in this paper was 7 pixels. The optimized cuts are shown by the green lines in \figref{fig:nucleiPartitioning}{b}.

  \section{Vertex-center cuts}
    The vertex-vertex cuts are not able to correctly partition clumps where the nuclei form triangles (the top three nuclei in \figref{fig:nucleiPartitioning}{}), and a new vertex must be introduced somewhere in the interior of the boundary. These type of cases are handled as follows: create triangles using the nuclei centers as vertices and add a new vertex to the center of each valid triangle, create cuts from center of the triangles to the boundary, and then optimize the cuts.

     \subsection{Triangulation}
      Using the nuclei center points, I construct a Delaunay triangulation, which has the property that if a circle is drawn through the three points of a triangle, then no other point is inside of the circle. This triangulation is shown in \figref{fig:nucleiPartitioning}{c}. Any triangle with an edge that intersects the boundary is removed; and, any triangle that has an interior angle larger than a threshold $\Theta_\mathrm{max}$ or smaller than a threshold $\Theta_\mathrm{min}$ is removed. Applying thresholds to the interior angles removes triangles where a new vertex should not be added (i.e.\ consider a clump of three nuclei that are close to co-linear, the triangle formed would have both a very large and two small interior angles). In this work $\Theta_\mathrm{min}=20^{\circ}$ and $\Theta_\mathrm{max}=110^{\circ}$.

     \subsection{Cut formation}
      The triangles are gathered into groups such that all triangles in a group share an edge with another triangle in the group. If a triangle is isolated, then it is the only triangle in its group. Iterate through each group and form cuts as follows
      \begin{enumerate}
        \item For each triangle in a group, and for each unshared edge in the triangle, find the boundary vertex $\vec{v}_p$ nearest to the midpoint of the unshared edge $\vec{m}$ and on the side of the edge opposite to the triangle center. Let $\uv{n}_m$ be the outward pointing unit vector of the unshared edge, then
        \begin{subequations}
          \begin{align}
          p &= \arg\min_{i\in\set{S}}\bigl(\lvert\vec{v}_i-\vec{m}\rvert\bigr) \\
          \set{S} &= \bigl\{s \st (\vec{v}_s-\vec{m})\cdot\uv{n}_{m}>0 \bigr\}.
          \end{align}
        \end{subequations}
        Create a cut from this vertex to the triangle center.
        \item Create cuts between the centers of two triangles that share an edge.
      \end{enumerate}
      The cuts formed following this method are shown as the red lines in \figref{fig:nucleiPartitioning}{d}.

      \begin{figure*}
        \centering
        \includegraphics[width=0.7\textwidth]{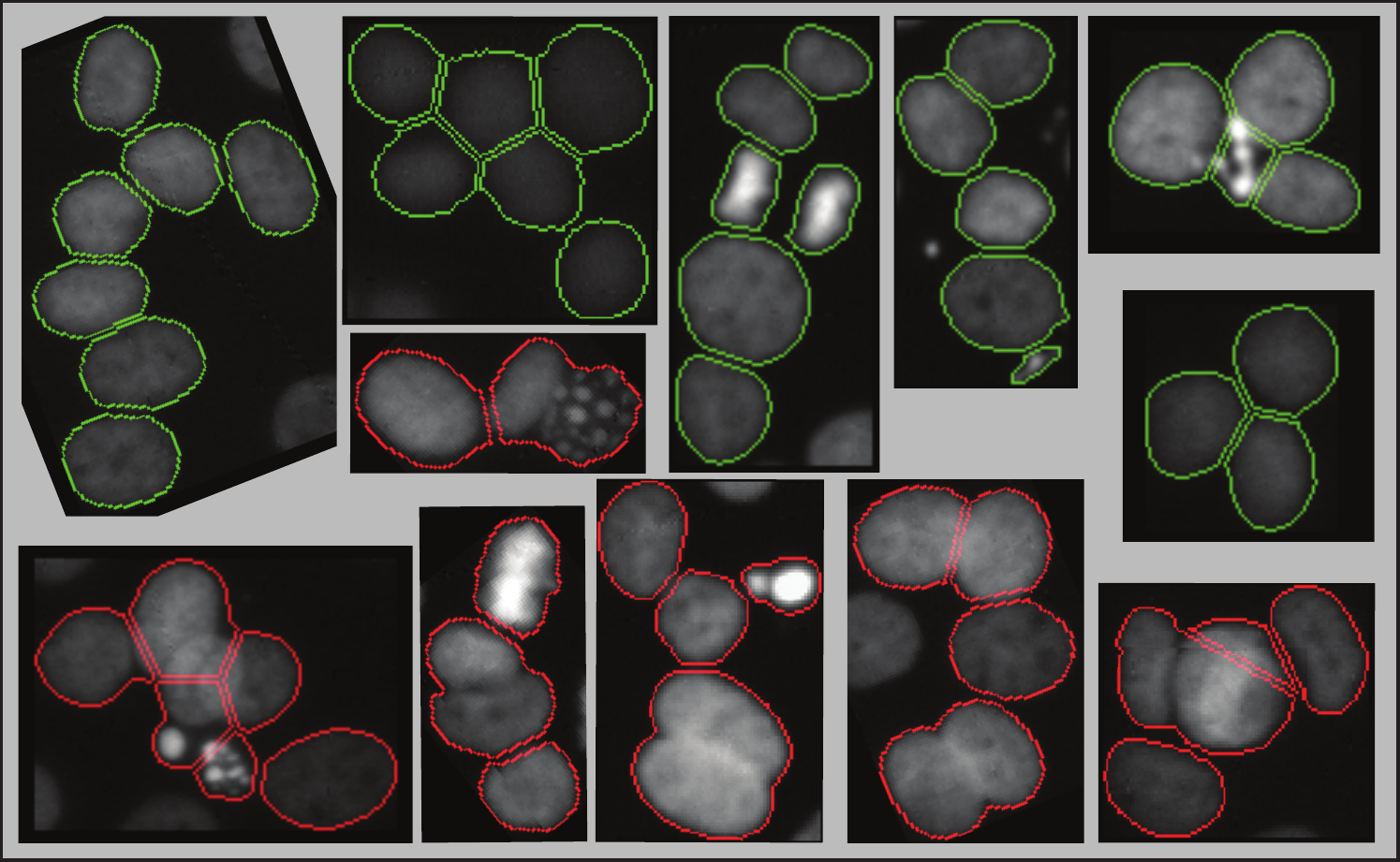}
        \caption{Partitioning results. Green/red lines show the final partitioning of 12 nuclei clumps. Green partition lines indicate correct partitioning, red partition lines indicate incorrect partitioning. (Note that incorrect partitioning can also be due to incorrect seed-point detection.)\label{fig:PartitioningResults}}
      \end{figure*}

     \subsection{Cut optimization}
      The vertex-center cuts are optimized in a similar manner to the vertex-vertex cuts, but the process is broken into two steps to reduce the search space: 1) optimize the cut vertices on the boundary, 2) optimize the location the added vertex (the triangle center).\par
      The boundary vertices are optimized one at a time by searching for a new vertex $\vec{v}_p$ in a small neighboring region around each original vertex that minimizes the distance to the triangle center $\vec{c}_t$, maximizes the dot-product between the boundary normal and the cut vector $\gv{\ell}_p = \vec{c}_t-\vec{v}_p$, and maximizes the curvature. That is, each cut vertex $\vec{v}_j$ connected to triangle center $\vec{c}_t$ is replaced with $\vec{v}_p$ where
      \begin{equation}
      p = \arg\max_{i\in\set{V}_j} \frac{\uv{n}_i\cdot\ugv{\ell}_{i}+\kappa_i}{|\gv{\ell}_i|}
      \end{equation}
      and where $\set{V}_j$ is the set of boundary indices for the neighboring region of $\vec{v}_j$. The geometry of this optimization is shown in the inset of \figref{fig:nucleiPartitioning}{d}.

      The center $\vec{c}_t$ of each triangle is optimized by looking for a new location $\vec{c}'_t$ in a small region $R$ around the triangle center (the shaded triangle in the inset of \figref{fig:nucleiPartitioning}{d}) such that the mean dot-product between the cuts and the boundary normals is maximized and the mean cut distance is minimized. Let $\set{V}$ be the set of boundary indices for the cut vertices connected to $\vec{c}_t$, and let $\set{M}$ be the set of triangle edge midpoints $\vec{m}$ for the shared edges of the triangle. The new center will be given by
      \begin{equation}
      \vec{c}'_t = \underset{\vec{x}\in\set{R}}{\arg\min} \frac{\sum_{i\in\set{V}} \uv{n}_{i}\cdot\ugv{\ell}_{xv_i} }{\sum_{i\in\set{V}}|\gv{\ell}_{xv_i}| + \sum_{\vec{m}\in\set{M}} |\vec{x}-\vec{m}|}
      \end{equation}
      where $\gv{\ell}_{xv_i} = \vec{x}-\vec{v}_i$ is a cut vector. The optimized cuts are shown by the greed lines in \figref{fig:nucleiPartitioning}{d}.

  \section{Cut selection}
    In regions where both vertex-vertex and vertex-center cuts exist (the top of the nuclei clump in \figref{fig:nucleiPartitioning}{}), the best set of cuts is chosen by voting for the set with the largest score in four categories: direction, curvature, overlap with image gradient, and overlap with inverted image. If a set wins three of the categories, then it is chosen; if both sets win two, then the set with the largest normalized cumulative score is chosen, and, if that is a tie, then the vertex-vertex cut set is chosen. The scores for each category are calculated as follows:

    \begin{enumerate}
      \item \emph{Direction}: The mean dot-product between boundary normals at the cut's boundary vertices and cut's direction for each set is used as the score. %For vertex-vertex cuts, let $\vec{v}_{j,(1,2)}$ and $\uv{n}_{j,(1,2)}$ be the two vertices and boundary normals for the $j$'th cut, then
    %	\begin{equation}
    %	A_{\mathrm{vv}}=\frac{1}{2N}\sum_{j=1}^N \bigl(\uv{n}_{j,1}\cdot\ugv{\ell}_{j}-\uv{n}_{j,2}\cdot\ugv{\ell}_{j}\bigr),
    %	\end{equation}
    %	where $\gv{\ell}_{j} = \vec{v}_{j,2}-\vec{v}_{j,1}$ and $N$ is the number of cuts. For the vertex-center cuts, let $\set{V}$ be the set of boundary indices of the cut vertices connected to the center $\vec{c}_t$,
    %	\begin{equation}
    %		A_{\mathrm{vc}} = \frac{1}{|\set{V}|}\sum_{i\in\set{V}} \frac{\uv{n}_i\cdot(\vec{c}_t-\vec{v}_i)}{|\vec{c}_t-\vec{v}_i|},
    %	\end{equation}
    %	where $|\set{V}|$ represents the number of elements in the set $\set{V}$. The normalized score is computed by divided each of these scores by the mean of the two.

      \item \emph{Curvature}: The mean boundary curvature for each set of cuts is used as the score. \emph{(When two vertex-vertex cuts have vertices that are one pixel apart, then the curvature at only one of these vertices is included in the mean calculation.)}

      \item \emph{Overlap with image gradient}: The mean value of the image gradient along the cuts for each set is used as the score. The image gradient (magnitude) is calculated as follows: smooth the original image with a Gaussian blur with $\sigma=1$, compute the gradient images $G_x$ and $G_y$ by convolving the derivative of a Gaussian with $\sigma=1$ along both image directions, compute the magnitude by $G=\sqrt{G_x^2+G_y^2}$, and finally smooth $G$ by morphologically closing using a disk with a radius of three pixels. An example of the gradient calculated in this way is shown in \figref{fig:nucleiPartitioning}{e}.

      \item \emph{Overlap with inverted image}: The mean value of the inverted image, $1/I$ where $I$ is the image intensity, along the cuts for each set is used as the score. The image is first smoothed with a Gaussian blur with $\sigma=1$. An example of the inverted image can be seen in \figref{fig:nucleiPartitioning}{e}.
    \end{enumerate}
    After computing the scores, the scores in each category are normalized by the mean score of the category. It is worth noting that the overlap with the image gradient, which gives the image edges, is used as most cuts should have a large overlap with the edges, and the overlap with the inverted image is used because, when there is a dark region between a group of nuclei, the image gradient magnitude will be small, but the inverted image will have a large value in such regions (see the top three nuclei in \figref{fig:nucleiPartitioning}{}). This allows for the correct set of cuts to be chosen, as can be seen in the final partitioning result shown in \figref{fig:nucleiPartitioning}{g}.

  \section{Validation}
    In order to validate this partitioning routine, I applied it to the same 2420 nuclei clumps as used in Ref. \onlinecite{2018arXiv180404071K}, and I used SALR clustering, from that same reference, to located the nuclei centers. I manually went through each clump and labeled the segmentation as being correct or incorrect, and the result was that $\sim93\%$ of the clumps were correctly partitioned. I show an example of six correct and six incorrectly partitioned clumps in \figref{fig:PartitioningResults}{}. There are three important things to note in these example clumps. 1) Three of the correctly partitioned clumps require 1, 2, or 3 new vertices to be added, thus, these clumps would not be partitioned correctly by any method that cannot add new vertices. 2) Blurriness does not effect the results. 3) Four of the incorrect partitions are only incorrect because two of the nuclei are still connected, while the other nuclei in the clump are correctly partitioned. This is a general trend the in the incorrectly segmented clumps; thus, many of the nuclei from the incorrectly segmented clumps, are correctly segmented.

    By means of comparison, I used CellProfiler \cite{carpenter2006cellprofiler}, which is a software package commonly used for image processing and analysis of biological images, to segmented the same 2420 nuclei clumps and manually labeled correct/incorrect segmentations. CellProfiler resulted in $\sim90\%$ of the clumps being correctly segmented, which means the geometric partitioning and SALR clustering performs better by $\sim3\%$.

  \section{Code and data availability}
    The code and validation data are available on GitHub at \url{https://github.com/jkpld/geometricPartitioning}.

  \section{Acknowledgment}
  I acknowledge support from the US Department of Energy Office of Science, Office of Basic Energy Sciences, under Award Number DE-FC02-04ER15533.

  \section*{References}

    \bibliography{bibliography}

\end{document}